\documentclass[runningheads]{llncs}
\usepackage[T1]{fontenc}
\usepackage{graphicx,verbatim}
\usepackage{amsfonts}
\usepackage{multirow}
\usepackage{booktabs}
\usepackage{subcaption}
\usepackage{amsmath} 
\usepackage{hyperref}

\begin{document}
\title{TemporalDoRA: Temporal PEFT for Robust Surgical Video Question Answering}

\titlerunning{TemporalDoRA}
%

\author{Luca Carlini\inst{1} \and
Chiara Lena\inst{1} \and
Cesare Hassan\inst{2} \and
Danail Stoyanov\inst{3} \and
Elena De Momi\inst{1} \and
Sophia Bano\inst{3} \and
Mobarak I. Hoque\inst{3,4}}

\authorrunning{Carlini et al.}

\institute{
Dipartimento di Elettronica, Informazione e Bioingegneria (DEIB), Politecnico di Milano, Italy\\
\email{luca.carlini@polimi.it}
\and
IRCCS Humanitas Research Hospital, Italy
\and
UCL Hawkes Institute and Department of Computer Science, University College London, UK
\and
Division of Informatics, Imaging and Data Science, University of Manchester, UK\\
\email{mobarak.hoque@manchester.ac.uk}
}
  
\maketitle              
\begin{abstract}
\sloppy


Surgical Video Question Answering (VideoQA) requires accurate temporal grounding while remaining robust to natural variation in how clinicians phrase questions, where linguistic bias can arise. Standard Parameter Efficient Fine Tuning (PEFT) methods adapt pretrained projections without explicitly modeling frame-to-frame interactions within the adaptation pathway, limiting their ability to exploit sparse temporal evidence. We introduce TemporalDoRA, a video-specific PEFT formulation that extends Weight-Decomposed Low-Rank Adaptation by (i) inserting lightweight temporal Multi-Head Attention (MHA) inside the low-rank bottleneck of the vision encoder and (ii) selectively applying weight decomposition only to the trainable low-rank branch rather than the full adapted weight. This design enables temporally-aware updates while preserving a frozen backbone and stable scaling. By mixing information across frames within the adaptation subspace, TemporalDoRA steers updates toward temporally consistent visual cues and improves robustness with minimal parameter overhead. To benchmark this setting, we present REAL-Colon-VQA, a colonoscopy VideoQA dataset with 6,424 clip--question pairs, including paired rephrased Out-of-Template questions to evaluate sensitivity to linguistic variation. TemporalDoRA improves Out-of-Template performance, and ablation studies confirm that temporal mixing inside the low-rank branch is the primary driver of these gains. We also validate on EndoVis18-VQA adapted to short clips and observe consistent improvements on the Out-of-Template split. Code and dataset available at~\href{https://anonymous.4open.science/r/TemporalDoRA-BFC8/}{Anonymous GitHub}.

\fussy
\end{abstract}

\section{Introduction}

During minimally invasive surgery, clinicians must read fast-changing endoscopic video to make high-stakes decisions where missed motion cues or transient findings can lead to complications \cite{khan2023current,Carlinigutjnl-2025-335091,massimi2025llmparis}. This demands models that integrate evidence across time, because many Video Question Answering (VideoQA) answers hinge on brief events such as tool actions, camera motion, and momentary occlusions \cite{surgicalVQA,Seenivasan2023SurgicalGPT,pitvqa,drago2025surgvivqa}. Although surgical Vision Language Models (VLMs) are improving, analyses show they remain largely text-centric during answer generation, with textual contributions outweighing visual ones \cite{parcalabescu2024vision}. This encourages reliance on linguistic priors rather than grounded visual evidence, leading to degraded performance under Out-of-Template rephrasing. In temporally rich surgical settings, such imbalance can obscure short-lived clinical events and reinforce shortcut learning from frequent answer patterns \cite{dong2024generalization,rosenfeld2025questioning,jiang2025knowing,wan2025eliminating,yuan2024advancing}.

Fully fine-tuning video VLMs is often impractical in clinical settings because it requires large annotated datasets, hard to obtain in the clinical field. Parameter Efficient Fine-Tuning (PEFT) addresses this by updating only a small subset of parameters~\cite{lora,zhang2023adaloraadaptivebudgetallocation,kopiczko2024veravectorbasedrandommatrix}. Low-Rank Adaptation (LoRA)~\cite{lora} learns a low-rank additive update, while Weight-Decomposed Low-Rank Adaptation (DoRA) factorizes the low-rank update into direction and per-channel magnitude improving scaling and optimization stability~\cite{liu2024doraweightdecomposedlowrankadaptation}. Video understanding typically needs time-aware architectures and prior work has shown benefits from explicit temporal modeling through aggregation, masked reconstruction, or sequence dynamics~\cite{xclip,videomae,videomamba}. ST-Adapter introduces temporal aggregation with a 3D convolutional adapter~\cite{pan2022stadapterparameterefficientimagetovideotransfer}. Unlike PEFT methods such as LoRA and DoRA, which apply low-rank updates directly to existing projection weights, adapter-based tuning inserts additional modules with a fixed bottleneck. This design can limit content adaptive integration and restrict update capacity when cross-modal alignment must shift under surgical domain change.


We propose TemporalDoRA, a video-specific PEFT method that extends DoRA in two complementary ways. First, we insert temporal Multi-Head Attention (MHA) inside the low-rank bottleneck of the vision encoder to enable frame-level interaction within the adaptation subspace. Second, we modify the post-bottleneck update by applying weight decomposition only to the trainable low-rank branch rather than reparameterizing the full adapted weight. This preserves the frozen backbone while enabling temporally-aware directional re-scaling. To evaluate robustness to linguistic variation, we introduce REAL-Colon-VQA, a colonoscopy VideoQA benchmark with paired In-Template and Out-of-Template questions. Across REAL-Colon-VQA and EndoVis18-VQA, TemporalDoRA improves Out-of-Template performance, indicating stronger robustness under rephrasing.
Our contributions are: 
\begin{itemize}

    \item TemporalDoRA: A video-PEFT formulation that introduces temporal MHA inside the low-rank bottleneck and selectively decomposes only the trainable low-rank branch. This enables temporally grounded adaptation while preserving backbone stability.

    \item REAL-Colon-VQA Dataset: A benchmark with 6,424 clip-question pairs designed to expose linguistic bias. With paired In-Template and Out-of-Template questions, it tests whether models ground answers in temporal evidence rather than memorized phrasing patterns.

    \item Robustness Analysis: An evaluation across two backbones showing that temporal mixing within the low-rank adaptation pathway is the main factor driving improved robustness to rephrasing.

\end{itemize}


\section{Methods}

We introduce TemporalDoRA, a video-specific PEFT formulation that modifies the low-rank adaptation pathway of the vision encoder to incorporate temporal mixing. Temporal modeling is applied only in the vision encoder, while the language model is adapted with standard DoRA, since the decoder operates on token sequences and receives temporally encoded visual inputs. We evaluate the approach on our colonoscopy dataset, REAL-Colon-VQA.

\subsection{TemporalDoRA}

\begin{figure}[t]
    \centering
    \includegraphics[width=0.9\linewidth]{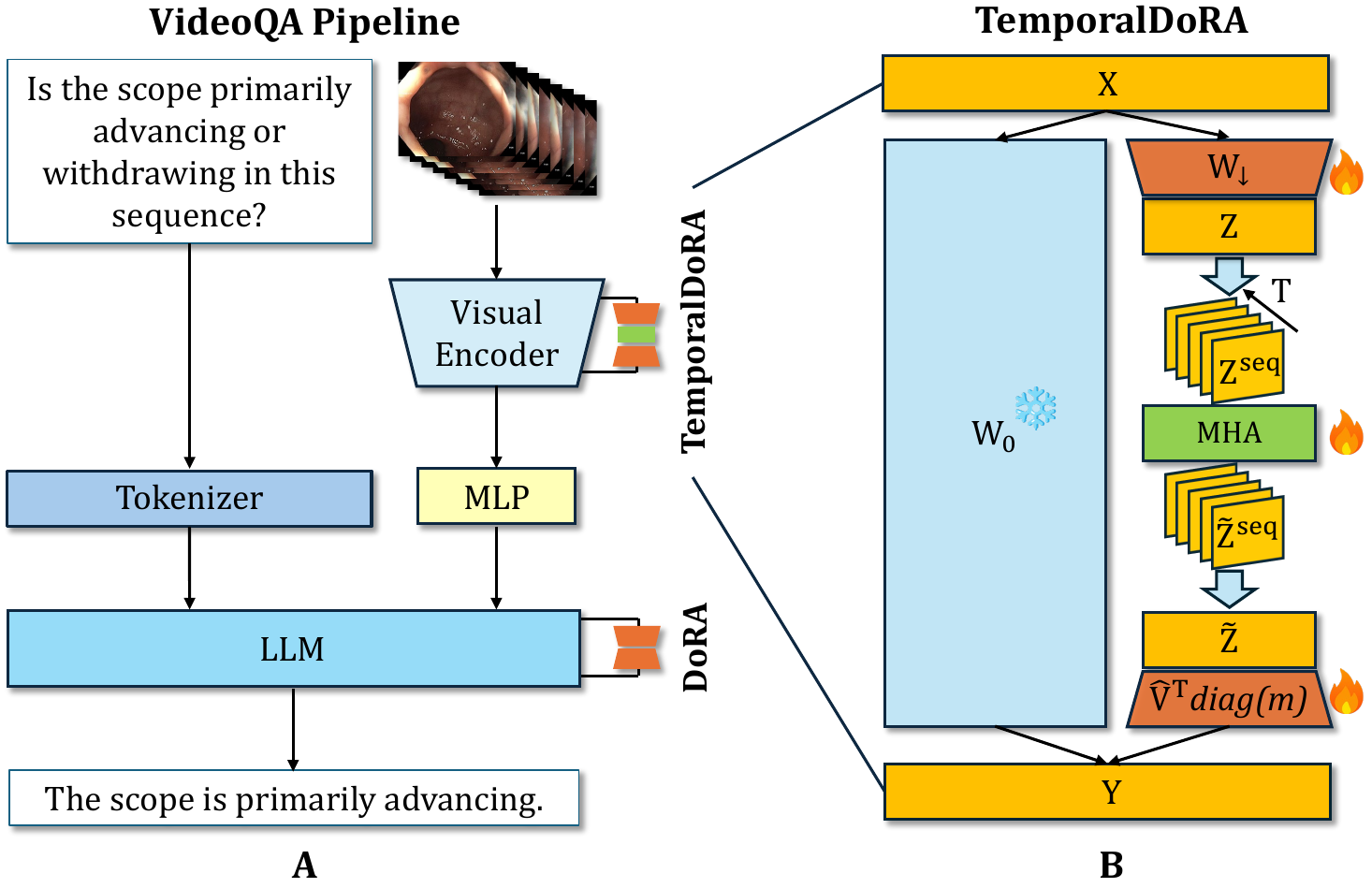}
    \caption{TemporalDoRA overview. (A) VideoQA pipeline. (B) TemporalDoRA: down-project to rank-$r$, apply temporal MHA over $T$, then selectively DoRA up-project and add to frozen $W_0$.}
    \label{fig:schema}
\end{figure}

TemporalDoRA extends DoRA for video understanding by inserting temporal mixing inside the low-rank adaptation bottleneck and modifying the up-projection parameterization. Fig.~\ref{fig:schema} summarizes the VideoQA pipeline and the TemporalDoRA module applied in the vision encoder. Let visual token features be
$X \in \mathbb{R}^{B \times T \times P \times C_{\text{in}}}$,
where $B$ is the batch size, $T$ the number of frames, $P$ the number of spatial tokens per frame, and $C_{\text{in}}$ the input channel dimension. Given a frozen projection
$W_0 \in \mathbb{R}^{C_{\text{in}} \times C_{\text{out}}}$,
the wrapped layer outputs $Y = XW_0 + \Delta$, where $\Delta$ is a trainable residual.

\noindent\textbf{Standard DoRA.}
DoRA reparameterizes the full effective weight as a normalized direction with a learnable per-output-channel magnitude, which improves scaling and optimization stability. The direction is computed from the sum of the frozen weight and the low-rank residual update, and the magnitude rescales this combined weight before applying it to the input features.

\noindent\textbf{Temporal MHA.}
Standard PEFT methods apply the low-rank update independently per token, so frames do not interact within the adaptation branch. We instead insert MHA inside the rank-$r$ bottleneck to mix information over time before up-projection. We use $4$ attention heads. We choose MHA because it performs content-dependent temporal aggregation where each frame can attend to the most informative frames while down-weighting redundant or corrupted ones, and multiple heads capture complementary temporal patterns. The down-projection compresses features:
\begin{equation}
Z = XW_{\downarrow},
\quad W_{\downarrow} \in \mathbb{R}^{C_{\text{in}} \times r},
\quad Z \in \mathbb{R}^{B \times T \times P \times r}.
\end{equation}
We reshape to per-location temporal sequences and apply attention along $T$:
\begin{equation}
Z^{\text{seq}} \in \mathbb{R}^{(B\cdot P) \times T \times r},
\quad
\widetilde{Z}^{\text{seq}} = \mathrm{MHA}(Z^{\text{seq}}),
\quad
\widetilde{Z} \in \mathbb{R}^{B \times T \times P \times r}.
\end{equation}
This enables content-dependent frame-to-frame aggregation within the adaptation subspace, allowing short-lived events to influence the update.

\noindent\textbf{Residual-only decomposition.}
Unlike standard DoRA, which applies direction--magnitude decomposition to the full effective weight \(W_0+\Delta W\), we apply it only to the trainable low-rank up-projection and keep \(W_0\) frozen. This choice preserves the pretrained directionality of the backbone and confines adaptation capacity to the low-rank branch, which is desirable in low-data surgical settings where full-weight reparameterization can overfit or distort well-calibrated features. In addition, applying the magnitude scaling after temporal mixing lets the model reweight output channels based on temporally aggregated evidence, while maintaining a stable initialization and a strictly zero-start residual.
We learn a direction matrix
$V \in \mathbb{R}^{C_{\text{out}} \times r}$ and normalize each row:
\begin{equation}
\widehat{V}(i,:) = \frac{V(i,:)}{\|V(i,:)\|_2 + \epsilon},
\end{equation}
together with a magnitude vector $m \in \mathbb{R}^{C_{\text{out}}}$:
\begin{equation}
W_{\uparrow} = \left(\mathrm{diag}(m)\,\widehat{V}\right)^{\top}
\in \mathbb{R}^{r \times C_{\text{out}}}.
\end{equation}
We initialize $m=0$ so the residual branch starts from zero output. 

\noindent\textbf{Residual pipeline.}
Using a scaling factor $\alpha$, the residual becomes
\begin{equation}
h(X) = XW_0 +
\frac{\alpha}{r}
\mathrm{MHA}\!\left(XW_{\downarrow}\right)\, W_{\uparrow}.
\end{equation}
We set $\alpha = 2r$ to maintain scale consistency across ranks. 

\noindent\textbf{Parameter efficiency.}
With the same backbone and tuned layers, ST-Adapter updates \(\approx 1.9\%\) of parameters, while TemporalDoRA updates \(\approx 0.22\%\) (\(\sim 8.6\times\) fewer).
We achieve this by adding only low-rank weights and applying DoRA only to the residual up-projection, keeping \(W_0\) frozen (no decomposition of \(W_0+\Delta W\)).

\subsection{REAL-Colon-VQA}\label{subsec:realcolonvqa}

\begin{figure}[t]
    \centering
    \includegraphics[width=\linewidth]{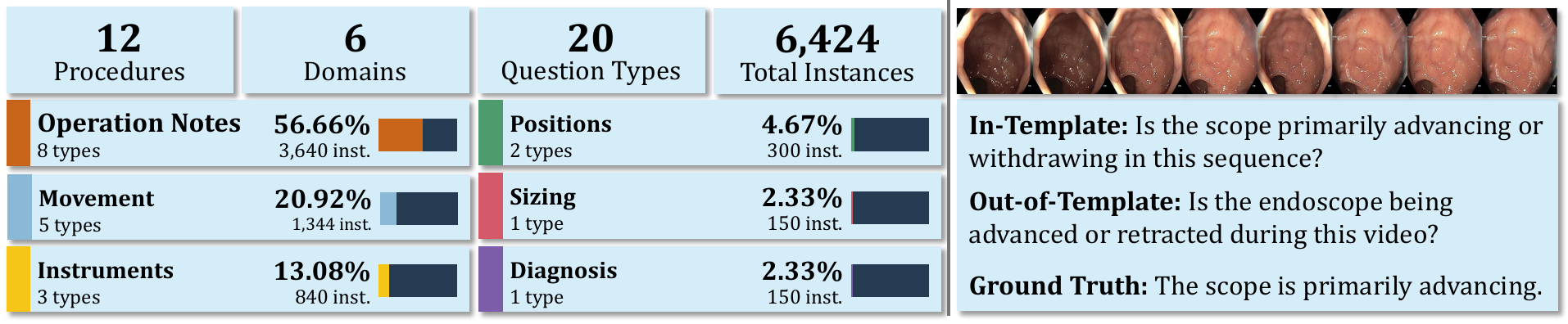}
    \caption{REAL-Colon-VQA question distribution and an example instance. }
    \label{fig:dataset}
\end{figure}

REAL-Colon-VQA (Fig.~\ref{fig:dataset}) is a colonoscopy benchmark for temporally grounded VideoQA built by extending REAL-Colon~\cite{biffi2024realcolon}. We annotate frame-level procedural dynamics relevant to temporal reasoning, including endoscope motion, tool usage, visibility/occlusions, flushing, and illumination mode. Lesion attributes  are inherited from REAL-Colon~\cite{biffi2024realcolon}. Motion, occlusion, and tool presence labels were produced by two trained non-clinical annotators with a third adjudicating disagreements.
We form 8-frame clips sampled from 30 fps videos with stride $=4$ (about $0.93$ s) and assign questions only when conditions hold for more than 5 frames to reduce single-frame cues. Out-of-Template paraphrased questions were initially generated using GPT-5.1 and subsequently reviewed and corrected by a human operator to ensure semantic equivalence and clinical consistency.

\section{Experimental setup and results}

\subsection{Implementation Details}

\noindent\textbf{Baselines.}
We implement LoRA~\cite{lora}, DoRA~\cite{liu2024doraweightdecomposedlowrankadaptation}, VeRA~\cite{kopiczko2024veravectorbasedrandommatrix}, and AdaLoRA~\cite{zhang2023adaloraadaptivebudgetallocation} using Hugging Face PEFT with rank \(r=8\).  
Following the same layer naming as TemporalDoRA, we adapt the vision encoder attention projections, namely the fused \(qkv\) input projection and the output projection, and we adapt the LLM decoder projections, namely \(q\), \(k\), \(v\), and the output projection.
For ST-Adapter~\cite{pan2022stadapterparameterefficientimagetovideotransfer}, we insert the temporal adapter branch into the same vision encoder attention projections using the original paper settings, and fine-tune the LLM decoder with DoRA for a fair comparison. We use two dataset-specific system prompts that encourage generalisation without enforcing rigid answer templates, unlike the longer prescriptive zero-shot prompts in~\cite{drago2025surgvivqa}. We evaluate PEFT methods on Qwen3-VL-2B~\cite{bai2025qwen3vltechnicalreport} and InternVL3-1B~\cite{zhu2025internvl3exploringadvancedtraining}. 
We additionally report SurgViVQA by running the authors' code on our train/val/test split, so absolute numbers may differ from~\cite{drago2025surgvivqa}.

\noindent\textbf{Datasets.}
We evaluate on REAL-Colon-VQA, our colonoscopy VideoQA benchmark built on REAL-Colon~\cite{biffi2024realcolon}, with 8-frame clips and paired In-Template questions and Out-of-Template rephrasings that act as an out-of-domain language shift, and on EndoVis18-VQA~\cite{surgicalVQA,qa_snne}, which we adapt to 8-frame clips aligned to the annotated frame and evaluate on the standard split with train sequences 2,3,4,6,7,9,10,11, validation sequences 12,14,15, and test sequences 1,5,16, including the perturbed Out-of-Template questions introduced in~\cite{qa_snne}.

\noindent\textbf{Training and evaluation.} 
Models are trained for 20 epochs with learning rate \(2 \times 10^{-4}\), selecting the best checkpoint by validation loss. We use batch size \(1\) with gradient accumulation \(8\). Experiments run on NVIDIA RTX 6000 Ada GPUs (SurgViVQA on H100). We report BLEU-4 (B4), ROUGE-L (RL), METEOR (MET), and keyword accuracy (Acc) following~\cite{drago2025surgvivqa}. In tables, \textbf{bold} indicates the best result and \underline{underlined} the second best.

\subsection{Results}

\begin{table*}[!t]
\caption{PEFT comparison on REAL-Colon-VQA and EndoVis18-VQA for In-Template and Out-of-Template splits using Qwen3-VL-2B~\cite{bai2025qwen3vltechnicalreport}, InternVL3-1B~\cite{zhu2025internvl3exploringadvancedtraining}, and SurgViVQA~\cite{drago2025surgvivqa}. \textbf{Bold} indicates best and \underline{underlined} second best within each model block and split.}
\label{tab:unified_three_blocks_peft_dataset_first_no_train}

\centering
\scriptsize
\setlength{\tabcolsep}{3.2pt}
\renewcommand{\arraystretch}{1.15}
\resizebox{\textwidth}{!}{%
\begin{tabular}{c c l | c c c c | c c c c}
\toprule
\multirow{2}{*}{\rotatebox[origin=c]{90}{\textbf{Data}}} &
\multirow{2}{*}{\rotatebox[origin=c]{90}{\textbf{Model}}} &
\multirow{2}{*}{\textbf{PEFT}} &
\multicolumn{4}{c|}{\textbf{In-Template}} &
\multicolumn{4}{c}{\textbf{Out-of-Template}} \\
\cmidrule(lr){4-7}\cmidrule(lr){8-11}
 &  &  & \textbf{B4} & \textbf{RL} & \textbf{MET} & \textbf{Acc} &
\textbf{B4} & \textbf{RL} & \textbf{MET} & \textbf{Acc} \\
\midrule

\multirow{15}{*}{\rotatebox[origin=c]{90}{\textbf{REAL-Colon-VQA}}}

& \multicolumn{2}{l|}{\textbf{SurgViVQA~\cite{drago2025surgvivqa}}}
& 0.649 & 0.812 & 0.815 & 0.472
& 0.142 & 0.375 & 0.379 & 0.199 \\
\cmidrule(lr){2-11}

& \multirow{7}{*}{\rotatebox[origin=c]{90}{\textbf{Qwen3-VL-2B}}}

& Zero-shot
& 0.196 & 0.514 & 0.538 & 0.457
& 0.112 & 0.433 & 0.459 & 0.470 \\

& & LoRA~\cite{lora}
& \textbf{0.742} & 0.815 & \underline{0.809} & 0.636
& 0.435 & 0.623 & 0.622 & 0.607 \\

& & DoRA~\cite{liu2024doraweightdecomposedlowrankadaptation}
& \underline{0.730} & \textbf{0.836} & \textbf{0.826} & \textbf{0.680}
& 0.246 & 0.611 & 0.617 & \textbf{0.669} \\

& & VeRA~\cite{kopiczko2024veravectorbasedrandommatrix}
& 0.249 & 0.524 & 0.539 & 0.501
& 0.138 & 0.436 & 0.463 & 0.504 \\

& & AdaLoRA~\cite{zhang2023adaloraadaptivebudgetallocation}
& 0.320 & 0.572 & 0.568 & 0.578
& 0.190 & 0.484 & 0.502 & 0.582 \\

& & ST-Adapter~\cite{pan2022stadapterparameterefficientimagetovideotransfer}
& 0.697 & 0.804 & 0.820 & 0.657
& \underline{0.466} & \underline{0.653} & \underline{0.680} & 0.622 \\

& & TemporalDoRA (ours)
& 0.703 & \underline{0.823} & 0.808 & \underline{0.678}
& \textbf{0.484} & \textbf{0.731} & \textbf{0.716} & \underline{0.646} \\

\cmidrule(lr){2-11}

& \multirow{7}{*}{\rotatebox[origin=c]{90}{\textbf{InternVL3-1B}}}

& Zero-shot
& 0.088 & 0.467 & 0.510 & 0.335
& 0.048 & 0.367 & 0.414 & 0.288 \\

& & LoRA~\cite{lora}
& 0.583 & 0.761 & 0.766 & 0.601
& 0.247 & 0.521 & 0.546 & \underline{0.527} \\

& & DoRA~\cite{liu2024doraweightdecomposedlowrankadaptation}
& \textbf{0.703} & \textbf{0.828} & \textbf{0.820} & \textbf{0.639}
& 0.212 & 0.530 & 0.565 & 0.508 \\

& & VeRA~\cite{kopiczko2024veravectorbasedrandommatrix}
& 0.238 & 0.570 & 0.580 & 0.413
& 0.128 & 0.446 & 0.488 & 0.387 \\

& & AdaLoRA~\cite{zhang2023adaloraadaptivebudgetallocation}
& 0.422 & 0.680 & 0.688 & 0.511
& 0.174 & 0.519 & 0.545 & 0.482 \\

& & ST-Adapter~\cite{pan2022stadapterparameterefficientimagetovideotransfer}
& 0.679 & 0.774 & 0.779 & 0.544
& \textbf{0.361} & \underline{0.569} & \underline{0.584} & 0.517 \\

& & TemporalDoRA (ours)
& \underline{0.609} & \underline{0.783} & \underline{0.777} & \underline{0.623}
& \underline{0.322} & \textbf{0.599} & \textbf{0.606} & \textbf{0.544} \\

\midrule

\multirow{9}{*}{\rotatebox[origin=c]{90}{\textbf{EndoVis18-VQA}}}


& \multicolumn{2}{l|}{\textbf{SurgViVQA~\cite{drago2025surgvivqa}}}
& 0.835 & 0.910 & 0.886 & 0.543
& 0.127 & 0.199 & 0.173 & 0.050 \\
\cmidrule(lr){2-11}

& \multirow{7}{*}{\rotatebox[origin=c]{90}{\textbf{Qwen3-VL-2B}}}
& Zero-shot
& 0.279 & 0.660 & 0.698 & 0.154
& 0.029 & 0.292 & 0.403 & 0.000 \\

& & LoRA~\cite{lora}
& 0.490 & 0.743 & \underline{0.875} & \underline{0.719}
& 0.116 & \underline{0.502} & 0.588 & \underline{0.304} \\

& & DoRA~\cite{liu2024doraweightdecomposedlowrankadaptation}
& 0.484 & 0.705 & 0.866 & \textbf{0.748}
& \underline{0.127} & 0.501 & \underline{0.619} & 0.276 \\

& & VeRA~\cite{kopiczko2024veravectorbasedrandommatrix}
& 0.465 & \textbf{0.824} & 0.853 & 0.480
& 0.014 & 0.270 & 0.395 & 0.025 \\

& & AdaLoRA~\cite{zhang2023adaloraadaptivebudgetallocation}
& \underline{0.568} & \underline{0.785} & 0.820 & 0.408
& 0.116 & 0.485 & 0.570 & 0.047 \\

& & ST-Adapter~\cite{pan2022stadapterparameterefficientimagetovideotransfer}
& 0.364 & 0.666 & 0.802 & 0.611
& 0.066 & 0.289 & 0.434 & 0.257 \\

& & TemporalDoRA (ours)
& \textbf{0.575} & 0.777 & \textbf{0.892} & 0.714
& \textbf{0.166} & \textbf{0.559} & \textbf{0.661} & \textbf{0.326} \\

\bottomrule
\end{tabular}%
}

\end{table*}

Table~\ref{tab:unified_three_blocks_peft_dataset_first_no_train} summarises the performance of the implemented PEFT methods across datasets and backbones. TemporalDoRA consistently improves robustness on the Out-of-Template split. On Qwen3-VL-2B for REAL-Colon-VQA, ROUGE-L increases to 0.731 compared with 0.653 for ST-Adapter, while maintaining competitive In-Template accuracy. On EndoVis18-VQA, TemporalDoRA achieves the strongest overall Out-of-Template performance, increasing keyword accuracy to 0.326 compared with 0.304 for LoRA. Table~\ref{tab:qual_examples_ood_qwen} shows that, compared to zero-shot, TemporalDoRA improves clinically grounded predictions while maintaining stability under rephrasing, whereas the zero-shot model often produces plausible but incorrect answers.

\noindent \textbf{Robustness to Rephrasing.}
TemporalDoRA is more stable than temporal-agnostic PEFT on Out-of-Template rephrasings, suggesting that temporal mixing helps predictions rely on temporally consistent visual evidence rather than phrasing-correlated shortcuts. By aggregating cues across frames inside the low-rank bottleneck, it reduces sensitivity to superficial wording changes and mitigates the language bias that emerges when temporal evidence is weakly encoded. 

\noindent\textbf{Why TemporalDoRA helps.}
TemporalDoRA encourages to rely on temporal video evidences across frames, reducing sensitivity to superficial phrasing variation and recovering short clinical cues distributed over time. ST-Adapter remains a strong baseline on Out-of-Template splits, underscoring the benefit of video-specific adaptation, but its adapter bottleneck  limit cross-modal realignment under surgical domain shift. SurgViVQA shows strong In-Template results but weaker Out-of-Template performance, which is consistent with higher sensitivity to question phrasing than to temporally grounded evidence.

\begin{table}[t]
\caption{Out-of-Template examples with Qwen3-VL-2B.}
\label{tab:qual_examples_ood_qwen}
\centering
\small
\setlength{\tabcolsep}{6pt}
\renewcommand{\arraystretch}{1.1}

\begin{tabular}{c p{0.74\textwidth}}
\toprule
\multicolumn{2}{l}{\textbf{REAL-Colon-VQA (Out-of-Template, Qwen3-VL-2B)}}\\
\midrule

\multirow{4}{*}{\raisebox{-0.5\height}{\includegraphics[width=0.20\textwidth,height=1.6cm,keepaspectratio]{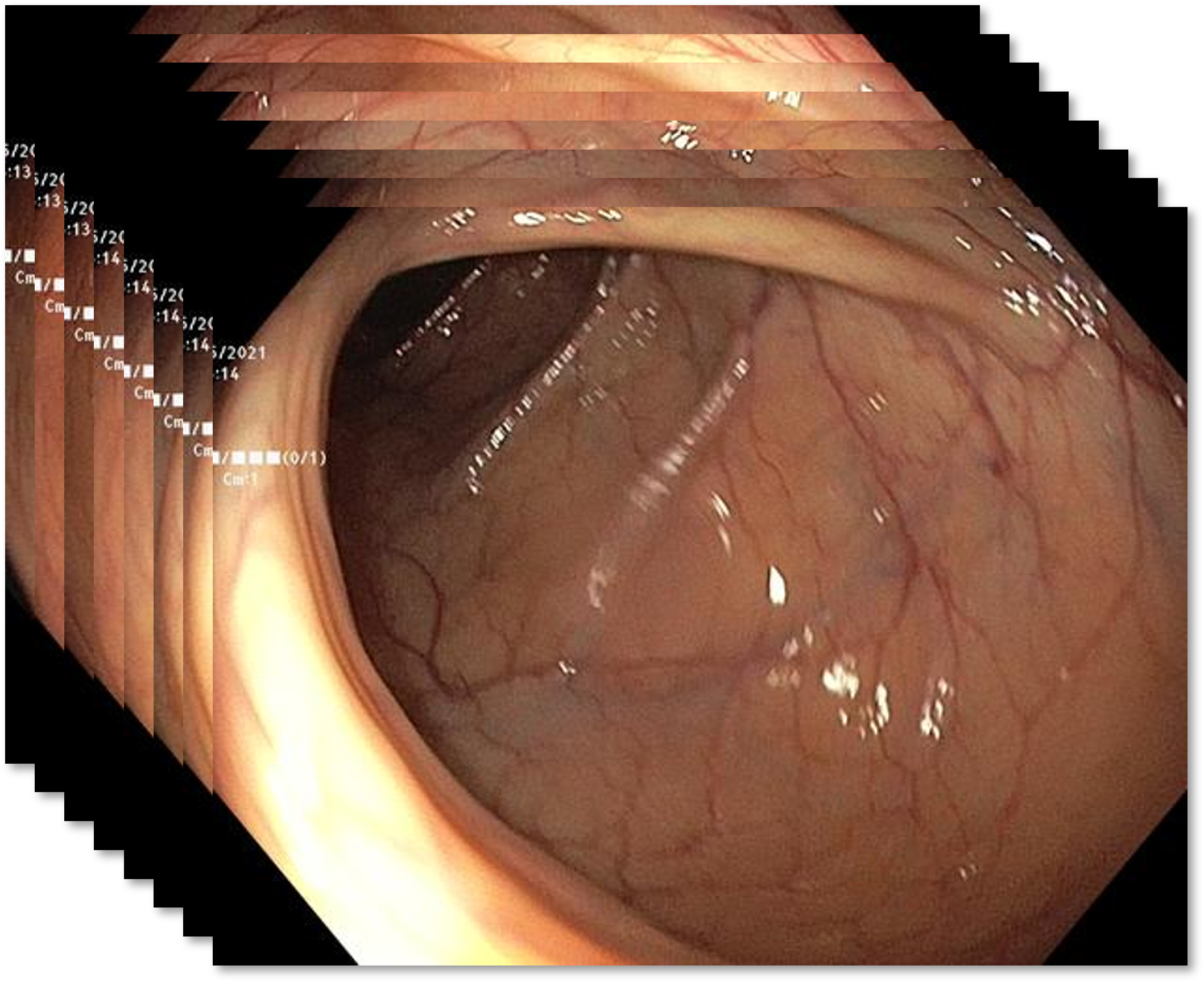}}}
& \textbf{Q1:} When motion occurs, does the scope pull backward?\\
& \textbf{GT:} No, the scope is advancing.\\
& \textbf{Pred (Zero-Shot):} No, the scope is advancing forward.\\
& \textbf{Pred (TemporalDoRA):} No, the scope is advancing.\\

\midrule
\multicolumn{2}{l}{\textbf{EndoVis18-VQA (Out-of-Template, Qwen3-VL-2B)}}\\
\midrule

\multirow{4}{*}{\raisebox{-0.5\height}{\includegraphics[width=0.20\textwidth,height=2cm,keepaspectratio]{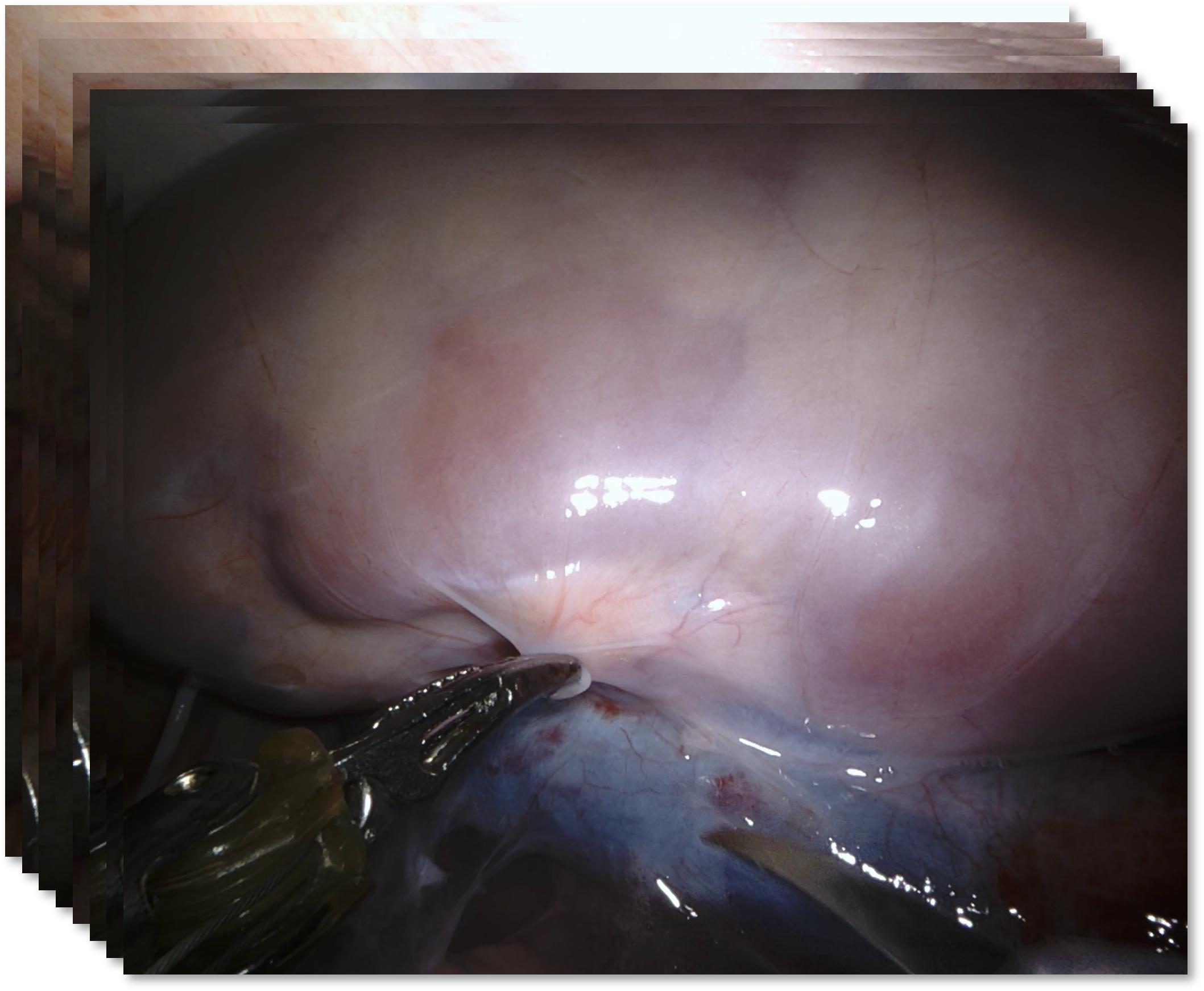}}}
& \textbf{Q2:} What specific abdominal organ is currently undergoing surgical intervention during the robotic-assisted procedure?\\
& \textbf{GT:} organ being operated is kidney\\
& \textbf{Pred (Zero-Shot):} organ being operated is liver\\
& \textbf{Pred (TemporalDoRA):} organ being operated is kidney\\

\bottomrule
\end{tabular}

\end{table}

\noindent\textbf{Ablation study.}
Table~\ref{tab:qwen_real_stdora_midlayer_ablation} compares temporal operators inserted inside the low-rank bottleneck of the vision encoder on REAL-Colon-VQA (Qwen3-VL-2B). MHA provides the best balance, particularly on Out-of-Template performance. LSTM improves In-Template metrics but reduces Out-of-Template generalization, and Mamba attains the highest Out-of-Template accuracy with lower text similarity, which may reflect the inherent temporal modeling capacity of these sequence-based operators. Table~\ref{tab:qwen_real_dora_lora_mhatt_singlecol_nomodel} evaluates inserting MHA in the low-rank bottleneck of LoRA and DoRA, showing consistent Out-of-Template gains over their baselines. It then compares TemporalDoRA to LoRA and DoRA, with and without MHA, isolating our second architectural change in which TemporalDoRA applies the DoRA direction--magnitude decomposition only to the low-rank residual branch while keeping the original weight frozen. TemporalDoRA achieves the strongest overall Out-of-Template results, indicating that combining bottleneck MHA with residual-only decomposition yields larger gains than adding MHA to existing PEFT methods alone.

\begin{table*}[t]
\caption{REAL-Colon-VQA (Qwen3-VL-2B). Ablation of the temporal operator in the low-rank bottleneck.}
\label{tab:qwen_real_stdora_midlayer_ablation}
\centering
\scriptsize
\setlength{\tabcolsep}{3.2pt}
\renewcommand{\arraystretch}{1.15}
\resizebox{\textwidth}{!}{%
\begin{tabular}{l | c c c c | c c c c}
\toprule
\multirow{2}{*}{\textbf{Temporal operator}} &
\multicolumn{4}{c|}{\textbf{In-Template}} &
\multicolumn{4}{c}{\textbf{Out-of-Template}} \\
\cmidrule(lr){2-5}\cmidrule(lr){6-9}
 & \textbf{B4} & \textbf{RL} & \textbf{MET} & \textbf{Acc} &
\textbf{B4} & \textbf{RL} & \textbf{MET} & \textbf{Acc} \\
\midrule

3D Convolution
& 0.618 & 0.781 & 0.761 & 0.647
& 0.312 & 0.606 & 0.616 & 0.609 \\

LSTM
& \textbf{0.717} & \textbf{0.828} & \textbf{0.818} & \underline{0.669}
& \underline{0.375} & \underline{0.653} & \underline{0.648} & 0.612 \\

Mamba
& 0.646 & 0.765 & 0.737 & 0.658
& 0.336 & 0.593 & 0.592 & \textbf{0.651} \\

Self-Attention
& 0.554 & 0.706 & 0.680 & 0.620
& 0.367 & 0.628 & 0.623 & 0.610 \\

MHA (Ours)
& \underline{0.703} & \underline{0.823} & \underline{0.808} & \textbf{0.678}
& \textbf{0.484} & \textbf{0.731} & \textbf{0.716} & \underline{0.646} \\

\bottomrule
\end{tabular}%
}

\end{table*}

\begin{table*}[t]
\caption{REAL-Colon-VQA (Qwen3-VL-2B). Effect of inserting MHA in the low-rank bottleneck of LoRA and DoRA compared to TemporalDoRA.}
\label{tab:qwen_real_dora_lora_mhatt_singlecol_nomodel}
\centering
\scriptsize
\setlength{\tabcolsep}{3.2pt}
\renewcommand{\arraystretch}{1.15}
\resizebox{\textwidth}{!}{%
\begin{tabular}{l | c c c c | c c c c}
\toprule
\multirow{2}{*}{\textbf{PEFT}} &
\multicolumn{4}{c|}{\textbf{In-Template}} &
\multicolumn{4}{c}{\textbf{Out-of-Template}} \\
\cmidrule(lr){2-5}\cmidrule(lr){6-9}
 & \textbf{B4} & \textbf{RL} & \textbf{MET} & \textbf{Acc} &
\textbf{B4} & \textbf{RL} & \textbf{MET} & \textbf{Acc} \\
\midrule

LoRA
& \textbf{0.742} & 0.814 & \underline{0.809} & 0.636
& 0.435 & 0.623 & 0.622 & 0.607 \\

LoRA+MHA
& 0.638 & 0.801 & 0.791 & \textbf{0.689}
& 0.305 & 0.627 & 0.642 & \underline{0.661} \\

\midrule

DoRA
& \underline{0.730} & \textbf{0.836} & \textbf{0.826} & \underline{0.680}
& 0.246 & 0.611 & 0.617 & \textbf{0.669} \\

DoRA+MHA
& 0.662 & 0.750 & 0.741 & 0.601
& \textbf{0.510} & \underline{0.649} & \underline{0.647} & 0.597 \\

\midrule

TemporalDoRA (Ours)
& 0.703 & \underline{0.823} & 0.808 & 0.678
& \underline{0.484} & \textbf{0.731} & \textbf{0.716} & 0.646 \\

\bottomrule
\end{tabular}%
}
\end{table*}




\section{Conclusion}

TemporalDoRA is a lightweight video PEFT method that injects temporal mixing into the low-rank adaptation pathway of the vision encoder retaining the optimization stability of weight decomposed updates. We also introduced REAL-Colon-VQA, a temporally grounded colonoscopy dataset with paired In-Template and Out-of-Template questions to measure robustness to rephrasing. Experiments on two backbones show that TemporalDoRA consistently improves Out-of-Template performance on both datasets, indicating more stable predictions under linguistic variation while maintaining competitive In-Template accuracy. The main limitation is that the within bottleneck MHA adds computation overhead and can become costly for long clips. Future work will focus on more efficient temporal operators and on extending PEFT into the LLM to further reduce language bias and improve robustness.

%
%
%
\bibliographystyle{splncs04}
\bibliography{mybibliography}

@inproceedings{pitvqa,
  title={PitVQA: Image-Grounded Text Embedding LLM for Visual Question Answering in Pituitary Surgery},
  author={He, Runlong and Xu, Mengya and Das, Adrito and Khan, Danyal Z. and Bano, Sophia and Marcus, Hani J. and Stoyanov, Danail and Clarkson, Matthew J. and Islam, Mobarakol},
  booktitle={Medical Image Computing and Computer Assisted Intervention, MICCAI 2024},
  pages={488--498},
  year={2024}
}

@inproceedings{Seenivasan2023SurgicalGPT,
  title={SurgicalGPT: End-to-End Language-Vision GPT for Visual Question Answering in Surgery},
  author={Seenivasan, Lalithkumar and Islam, Mobarakol and Kannan, Gokul and Ren, Hongliang},
  booktitle={Medical Image Computing and Computer Assisted Intervention, MICCAI 2023},
  pages={283--293},
  year={2023}
}

@inproceedings{videomae,
  title={VideoMAE: Masked Autoencoders are Data-Efficient Learners for Self-Supervised Video Pre-Training},
  author={Tong, Zhan and Song, Yibing and Wang, Jue and Wang, Limin},
  booktitle={Advances in Neural Information Processing Systems},
  year={2022}
}

@inproceedings{lora,
  title={LoRA: Low-Rank Adaptation of Large Language Models},
  author={Hu, Edward J. and Shen, Yelong and Wallis, Phillip and Allen-Zhu, Zeyuan and Li, Yuanzhi and Wang, Shean and Wang, Lu and Chen, Weizhu},
  booktitle={International Conference on Learning Representations},
  year={2022}
}

@inproceedings{surgicalVQA,
  title={Surgical-VQA: Visual Question Answering in Surgical Scenes Using Transformer},
  author={Seenivasan, Lalithkumar and Islam, Mobarakol and Krishna, Adithya K. and Ren, Hongliang},
  booktitle={Medical Image Computing and Computer Assisted Intervention, MICCAI 2022},
  pages={33--43},
  year={2022}
}

@inproceedings{xclip,
  title={X-CLIP: End-to-End Multi-grained Contrastive Learning for Video-Text Retrieval},
  author={Ma, Yiwei and Xu, Guohai and Sun, Xiaoshuai and Yan, Ming and Zhang, Ji and Ji, Rongrong},
  booktitle={Proceedings of the 30th ACM International Conference on Multimedia},
  pages={638--647},
  year={2022}
}

@inproceedings{videomamba,
  title={VideoMamba: State Space Model for Efficient Video Understanding},
  author={Li, Kunchang and Li, Xinhao and Wang, Yi and He, Yinan and Wang, Yali and Wang, Limin and Qiao, Yu},
  booktitle={Computer Vision, ECCV 2024},
  pages={237--255},
  year={2025}
}

@article{biffi2024realcolon,
  title={REAL-Colon: A dataset for developing real-world AI applications in colonoscopy},
  author={Biffi, Carlo and Antonelli, Giulio and Bernhofer, Sebastian and Hassan, Cesare and Hirata, Daizen and Iwatate, Mineo and Maieron, Andreas and Salvagnini, Pietro and Cherubini, Andrea},
  journal={Scientific Data},
  volume={11},
  number={539},
  year={2024}
}

@article{Carlinigutjnl-2025-335091,
  title={Large language models for detecting colorectal polyps in endoscopic images},
  author={Carlini, Luca and Massimi, Davide and Mori, Yuichi and Antonelli, Giulio and Rizkala, Tommy and Spadaccini, Marco and Lena, Chiara and Parasa, Sravanthi and Bisschops, Raf and von Renteln, Daniel and O{\textquoteright}Reilly, Susanne and Sharma, Prateek and Rex, Douglas K. and Bretthauer, Michael and Repici, Alessandro and De Momi, Elena and Hassan, Cesare},
  journal={Gut},
  year={2025}
}

@article{massimi2025llmparis,
  title={Large language model for interpreting the Paris classification of colorectal polyps},
  author={Massimi, Davide and Carlini, Luca and Mori, Yuichi and Di Stefano, Luca and Antonelli, Giulio and Rizkala, Tommy and Spadaccini, Marco and de Sire, Roberto and Alfarone, Ludovico and Lena, Chiara and others},
  journal={Endoscopy International Open},
  volume={13},
  number={continuous publication},
  year={2025},
  publisher={Georg Thieme Verlag KG}
}

@article{pan2022stadapterparameterefficientimagetovideotransfer,
  title={St-adapter: Parameter-efficient image-to-video transfer learning},
  author={Pan, Junting and Lin, Ziyi and Zhu, Xiatian and Shao, Jing and Li, Hongsheng},
  journal={Advances in Neural Information Processing Systems},
  volume={35},
  pages={26462--26477},
  year={2022}
}

@inproceedings{liu2024doraweightdecomposedlowrankadaptation,
  title={Dora: Weight-decomposed low-rank adaptation},
  author={Liu, Shih-Yang and Wang, Chien-Yi and Yin, Hongxu and Molchanov, Pavlo and Wang, Yu-Chiang Frank and Cheng, Kwang-Ting and Chen, Min-Hung},
  booktitle={Forty-first International Conference on Machine Learning},
  year={2024}
}

@article{bai2025qwen3vltechnicalreport,
  title={Qwen3-VL Technical Report},
  author={Bai, Shuai and Cai, Yuxuan and Chen, Ruizhe and Chen, Keqin and Chen, Xionghui and others},
  journal={arXiv preprint arXiv:2511.21631},
  year={2025}
}

@article{zhu2025internvl3exploringadvancedtraining,
  title={Internvl3: Exploring advanced training and test-time recipes for open-source multimodal models},
  author={Zhu, Jinguo and Wang, Weiyun and Chen, Zhe and Liu, Zhaoyang and Ye, Shenglong and Gu, Lixin and Tian, Hao and Duan, Yuchen and Su, Weijie and Shao, Jie and others},
  journal={arXiv preprint arXiv:2504.10479},
  year={2025}
}

@article{kopiczko2024veravectorbasedrandommatrix,
  title={VeRA: Vector-based Random Matrix Adaptation},
  author={Dawid J. Kopiczko and Tijmen Blankevoort and Yuki Markus Asano},
  journal={ArXiv},
  year={2023},
  volume={abs/2310.11454},
  url={https://api.semanticscholar.org/CorpusID:264172315}
}

@article{drago2025surgvivqa,
  title={SurgViVQA: Temporally-Grounded Video Question Answering for Surgical Scene Understanding},
  author={Drago, Mauro Orazio and Carlini, Luca and Balyemez, Pelinsu Celebi and Pierantozzi, Dennis and Lena, Chiara and Hassan, Cesare and Stoyanov, Danail and De Momi, Elena and Bano, Sophia and Hoque, Mobarak I},
  journal={arXiv preprint arXiv:2511.03325},
  year={2025}
}

@article{zhang2023adaloraadaptivebudgetallocation,
  title={Adalora: Adaptive budget allocation for parameter-efficient fine-tuning},
  author={Zhang, Qingru and Chen, Minshuo and Bukharin, Alexander and Karampatziakis, Nikos and He, Pengcheng and Cheng, Yu and Chen, Weizhu and Zhao, Tuo},
  journal={arXiv preprint arXiv:2303.10512},
  year={2023}
}

@article{qa_snne,
  title={When to Trust the Answer: Question-Aligned Semantic Nearest Neighbor Entropy for Safer Surgical VQA},
  author={Pierantozzi, Dennis and Carlini, Luca and Drago, Mauro Orazio and Lena, Chiara and Hassan, Cesare and De Momi, Elena and Stoyanov, Danail and Bano, Sophia and Hoque, Mobarak I},
  journal={arXiv preprint arXiv:2511.01458},
  year={2025}
}

@article{yuan2024advancing,
  title={Advancing surgical vqa with scene graph knowledge},
  author={Yuan, Kun and Kattel, Manasi and Lavanchy, Joel L and Navab, Nassir and Srivastav, Vinkle and Padoy, Nicolas},
  journal={International journal of computer assisted radiology and surgery},
  volume={19},
  number={7},
  pages={1409--1417},
  year={2024},
  publisher={Springer}
}

@article{dong2024generalization,
  title={Generalization or memorization: Data contamination and trustworthy evaluation for large language models},
  author={Dong, Yihong and Jiang, Xue and Liu, Huanyu and Jin, Zhi and Gu, Bin and Yang, Mengfei and Li, Ge},
  journal={arXiv preprint arXiv:2402.15938},
  year={2024}
}

@inproceedings{wan2025eliminating,
  title={Eliminating Language Bias for Medical Visual Question Answering with Counterfactual Contrastive Training},
  author={Wan, Xingyu and Teng, Qiaoying and Chen, Jun and Lu, Yonghan and Yuan, Deqi and Liu, Zhe},
  booktitle={International Conference on Medical Image Computing and Computer-Assisted Intervention},
  pages={194--204},
  year={2025},
  organization={Springer}
}

@inproceedings{jiang2025knowing,
  title={Knowing or Guessing? Robust Medical Visual Question Answering via Joint Consistency and Contrastive Learning},
  author={Jiang, Songtao and Chen, Yuxi and Song, Sibo and Zhang, Yan and Jin, Yeying and Feng, Yang and Wu, Jian and Liu, Zuozhu},
  booktitle={International Conference on Medical Image Computing and Computer-Assisted Intervention},
  pages={325--335},
  year={2025},
  organization={Springer}
}

@article{rosenfeld2025questioning,
  title={Questioning the Stability of Visual Question Answering},
  author={Rosenfeld, Amir and Glazer, Neta and Fetaya, Ethan},
  journal={arXiv preprint arXiv:2511.11206},
  year={2025}
}

@article{parcalabescu2024vision,
  title={Do Vision \& Language Decoders use Images and Text equally? How Self-consistent are their Explanations?},
  author={Parcalabescu, Letitia and Frank, Anette},
  journal={arXiv preprint arXiv:2404.18624},
  year={2024}
}

@article{khan2023current,
  title={Current and future advances in surgical therapy for pituitary adenoma},
  author={Khan, Danyal Z and Hanrahan, John G and Baldeweg, Stephanie E and Dorward, Neil L and Stoyanov, Danail and Marcus, Hani J},
  journal={Endocrine Reviews},
  volume={44},
  number={5},
  pages={947--959},
  year={2023},
  publisher={Oxford University Press US}
}

\end{document}